\documentclass{ecai}

\usepackage{latexsym}
\usepackage{amssymb}
\usepackage{amsmath}
\usepackage{amsthm}
\usepackage{booktabs}
\usepackage{enumitem}
\usepackage{graphicx}
\usepackage{color}

\usepackage[british]{babel}
\usepackage[hidelinks]{hyperref}
\usepackage[capitalize]{cleveref}
\usepackage{mathtools}
\usepackage{booktabs}
\usepackage{fancyhdr}
\usepackage[shortcuts]{extdash}
\usepackage[all]{foreign}
\usepackage{bm}

\usepackage{tikz}
\usetikzlibrary{arrows.meta, positioning}

\newcommand{\BibTeX}{B\kern-.05em{\sc i\kern-.025em b}\kern-.08em\TeX}

\usepackage{amsmath}        %
\usepackage{amssymb}        %
\usepackage{bm}             %
\usepackage{graphicx}       %
\usepackage{subfig}         %
\usepackage{booktabs}       %
\usepackage{multirow}       %
\usepackage{hyperref}       %
\usepackage{cleveref}       %
\usepackage{wrapfig}        %
\usepackage{algorithm}      %
\usepackage{algpseudocode}  %
\usepackage{float}          %
\usepackage{xcolor}         %
\usepackage{natbib}         %
\usepackage{url}            %
\usepackage{mathtools}      %

\begin{document}

\begin{frontmatter}

\paperid{123}

\title{Methodological Insights into Structural Causal Modelling and Uncertainty-Aware Forecasting for Economic Indicators}

\author[A,B,C,D]{\fnms{Federico}~\snm{Cerutti}\orcid{0000-0003-0755-0358}\thanks{Corresponding Author. Email: federico.cerutti@unibs.it.}}

\address[A]{University of Brescia, Italy}
\address[B]{Imperial College London, UK}
\address[C]{Cardiff University, UK}
\address[D]{University of Southampton, UK}

\begin{abstract}
This paper presents a methodological approach to financial time series analysis by combining causal discovery and uncertainty-aware forecasting. As a case study, we focus on four key U.S. macroeconomic indicators \--- GDP, economic growth, inflation, and unemployment \--- and we apply the LPCMCI framework with Gaussian Process Distance Correlation (GPDC) to uncover dynamic causal relationships in quarterly data from 1970 to 2021. Our results reveal a robust unidirectional causal link from economic growth to GDP and highlight the limited connectivity of inflation, suggesting the influence of latent factors. Unemployment exhibits strong autoregressive dependence, motivating its use as a case study for probabilistic forecasting. Leveraging the Chronos framework, a large language model trained for time series, we perform zero-shot predictions on unemployment. This approach delivers accurate forecasts one and two quarters ahead, without requiring task-specific training. Crucially, the model's uncertainty-aware predictions yield 90\%  confidence intervals, enabling effective anomaly detection through statistically principled deviation analysis. This study demonstrates the value of combining causal structure learning with probabilistic language models to inform economic policy and enhance forecasting robustness.
\end{abstract}

\end{frontmatter}

\section{Introduction}
Time series analysis is foundational in financial modelling, supporting tasks ranging from risk management and portfolio optimisation to macroeconomic forecasting. In multivariate financial settings \--- such as analysing interest rates, exchange rates, asset prices, and credit indicators simultaneously \--- uncovering the causal relationships among variables is crucial for both interpretability and intervention. Structural causal models for time series, such as the LPCMCI framework (see \Cref{sec:lpcmci}), enable the identification of dynamic and contemporaneous causal links through conditional independence testing, offering a principled approach to disentangle complex interdependencies in financial systems. On the other hand, forecasting a single financial variable \--- such as a stock price, inflation rate, or cryptocurrency valuation \--- under uncertainty is equally vital. Recent advances in large pre-trained models based on language model architectures, such as \texttt{Chronos} \cite{ansari2024chronos} (see \Cref{sec:Chronos}),\footnote{\url{https://github.com/amazon-science/chronos-forecasting} (on 23 May 2025).} have enabled the generation of probabilistic forecasts that capture the full distribution of possible outcomes. This is particularly valuable in finance, where understanding the range and likelihood of future scenarios directly informs trading strategies, risk exposure assessments, and regulatory compliance.

In this study, we analyse the causal interplay among four key macroeconomic indicators in the United States \--- Gross Domestic Product (GDP), economic growth, inflation, and unemployment \--- over the period 1970 to 2021. Consistent with the methodology proposed by \cite{huang2019causal}, we employ quarterly aggregated data, normalised via standard scaling to ensure numerical stability and comparability across variables. 
The resulting causal graph (see \Cref{sec:resultsgpdc}) highlights a robust unidirectional link from Economic Growth to GDP, with no evidence of the reverse, while Inflation remains weakly connected and potentially influenced by latent factors. In contrast, Unemployment exhibits strong autoregressive structure, justifying its selection for targeted probabilistic forecasting.

Given the pronounced temporal self-dependence observed in the Unemployment series \--- characterised by robust causal links to its own lagged values across consecutive quarters \--- we focus the subsequent analysis on this variable. Specifically, we investigate the predictive structure of Unemployment using a zero-shot probabilistic forecasting approach based on a large pretrained language model.

This methodological choice offers two key advantages: zero-shot capabilities allow the model to generalise patterns without any task-specific training, thus enabling rapid deployment in dynamic settings; and uncertainty-awareness provides predictive intervals critical for reliable decision-making. Notably, our results demonstrate remarkable predictive accuracy at one and two time steps ahead, even without model fine-tuning. Furthermore, the probabilistic nature of the forecasts enables anomaly detection through interval-based reasoning: when observed values fall outside the predicted 90\% confidence interval, this deviation is unlikely under the model's assumptions, suggesting the presence of structural shifts, regime changes, or exogenous shocks. Such out-of-bound observations can thus serve as rigorous flags for policymakers or analysts.

\section{Background}

\subsection{Causal analysis using LPCMCI}
\label{sec:lpcmci}

PCMCI \cite{runge_Detectingquantifying_19} is a constraint-based causal discovery method designed to estimate causal networks from highly-interdependent large-scale time series by accurately identifying the best variables to condition on. It consists of two parts:
\begin{itemize}
    \item First, the condition selection part of the algorithm is performed via PC$_1$ (a variant of the PC algorithm \cite{spirtes_AlgorithmFastRecovery_91}) to identify the relevant conditioning sets for all time series variables amongst the parents of each variable;
    \item Second, a Momentary Conditional Independence (MCI) test is performed to test whether two variables are independent, given their parent sets.
\end{itemize}

The two stages of the PCMCI algorithm serve two purposes: the first one removes irrelevant conditions for each variable by iterative independence testing, while the second one addresses false-positive control for the highly correlated time series the algorithm can consider.
PCMCI rests on the assumptions of causal sufficiency, the causal Markov condition, faithfulness, the absence of contemporaneous causal effects and stationarity.

Latent PCMCI (LPCMCI) \cite{gerhardus2020high} is a constraint-based causal discovery algorithm designed to extend the PCMCI framework to settings with latent confounders. While PCMCI improves statistical power by adaptively refining the conditioning sets in its MCI step, LPCMCI further constrains these sets by restricting them to known ancestors of the tested variables and systematically augmenting them with default conditions based on identified parents. This design aims to account for hidden common causes and to stabilise conditional independence testing in the presence of latent variables.

A key innovation of LPCMCI is the introduction of middle marks in its graphical representation, which provide intermediate causal information during the algorithm’s execution. These marks allow for early and precise edge orientations by encoding partial ancestor–descendant relationships before determining the final graph. Unlike previous methods that separate edge removal and orientation into distinct phases, LPCMCI integrates these steps, ensuring the evolving graph remains interpretable at all stages.

LPCMCI proceeds iteratively, starting from a fully connected graph and progressively refining it. During an initial phase, many false links are removed through CI tests that leverage the increased effect size from default conditioning. Simultaneously, middle marks guide early orientation decisions, further refining the structure. The remaining erroneous edges are pruned in the final phase, and orientations are finalized using additional statistical tests and refined graphical rules.

Under standard assumptions of faithfulness and sufficient sample size, LPCMCI is sound and complete, meaning it reliably recovers the true Partial Ancestral Graph (PAG). Its ability to integrate latent confounders while improving recall and reducing false positives makes it particularly suited for high-dimensional and time-series data.

Our study used LPCMCI mainly in conjunction with a specific MCI test, the Gaussian Process Distance Correlation (GPDC) test \cite{szekely2007measuring}, which is a nonparametric framework for testing conditional independence \( X \perp\!\!\!\perp Y \mid Z \) in continuous-valued settings. The approach decomposes the problem into two stages: regression followed by dependence testing. Let \( (X, Y, Z) \in \mathbb{R}^{d_X} \times \mathbb{R}^{d_Y} \times \mathbb{R}^{d_Z} \) be random variables with unknown joint distribution. The null hypothesis assumes that, conditional on \( Z \), the variables \( X \) and \( Y \) are independent. Under this hypothesis, one can model the conditional expectations via smooth functions:
\begin{equation}
    X = f_X(Z) + \varepsilon_X, \qquad Y = f_Y(Z) + \varepsilon_Y,
\end{equation}
where \( \varepsilon_X, \varepsilon_Y \) are residual random variables satisfying \( \mathbb{E}[\varepsilon_X \mid Z] = \mathbb{E}[\varepsilon_Y \mid Z] = 0 \). If the null holds, the residuals \( \varepsilon_X \) and \( \varepsilon_Y \) should be mutually independent.

The core idea is to estimate \( f_X \) and \( f_Y \) nonparametrically using Gaussian process regression, which permits flexible modelling of complex functional dependencies without committing to a specific parametric form. Once the regression functions are fitted, the empirical residuals are computed. Let \( \hat{\varepsilon}_X \) and \( \hat{\varepsilon}_Y \) denote the estimated residuals after regressing out \( Z \). These are then marginally transformed to obtain variables \( r_X \) and \( r_Y \) with uniform marginals via rank-based normalisation, a standard step to ensure the dependence test is sensitive to a wide range of relationships. The test statistic is defined as the distance correlation
\begin{equation}
\mathcal{R}(r_X, r_Y) = \frac{\mathcal{V}^2(r_X, r_Y)}{\sqrt{\mathcal{V}^2(r_X, r_X) \, \mathcal{V}^2(r_Y, r_Y)}},
\end{equation}
where \( \mathcal{V}^2(\cdot, \cdot) \) denotes the squared population distance covariance. At the population level, \(\mathcal{R}(r_X, r_Y) = 0\) if and only if \( r_X \) and \( r_Y \) are independent, which establishes the theoretical consistency of the test. For the sample version, this property holds asymptotically under suitable regularity conditions, such as ergodicity or sufficiently strong mixing, which ensure that empirical averages converge to their population counterparts.

The statistical validity of the GPDC procedure hinges on the asymptotic behaviour of the distance correlation under the null. However, unlike parametric tests, the null distribution of \( \mathcal{R}(r_X, r_Y) \) is not analytically tractable in this setting due to the presence of estimated residuals and the nonlinearity of the transformations involved. As a result, the significance of the observed statistic is assessed via a permutation or resampling-based approximation of the null distribution. Although the GPDC framework is theoretically appealing, as it combines the expressive power of kernel-based regression with the rigorous sensitivity of distance correlation, its practical application is subject to limitations arising from high-dimensional conditioning. In particular, when the size of the conditioning sets is large relative to the available sample size, the Gaussian Process regression may fail to capture conditional structures adequately, leaving the test close to an unconditional dependence measure. We will investigate further these aspects in future work.

\begin{figure*}[bt]
    \centering
    \includegraphics[width=\linewidth]{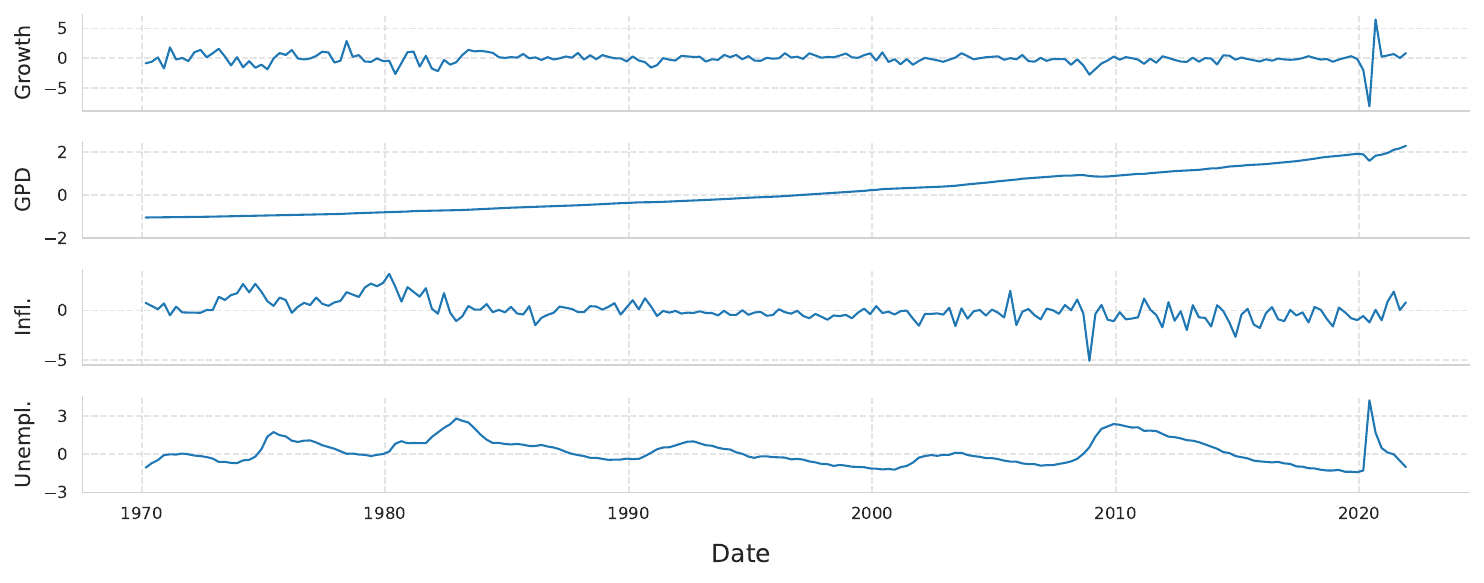}
    \caption{Visualisation of the normalised time series analyised.}
    \label{fig:time-series}
    \vspace{1em}
\end{figure*}

\subsection{LLM-based Time Series for Zero Shot Forecasting: Chronos}
\label{sec:Chronos}

Recent work has proposed adapting large-scale language models to the domain of probabilistic time series forecasting by redefining the input representation and training paradigm in a way that preserves architectural compatibility \cite{ansari2024chronos}. The proposed approach, referred to as \texttt{Chronos}, leverages the sequential modelling capabilities of transformer-based language models, such as T5~\cite{raffel2020exploring} and GPT-2~\cite{radford2019language}, by introducing a tokenisation scheme that maps real-valued time series into a discrete token space. This enables direct reuse of pretrained language model architectures without altering the core attention mechanisms or optimisation procedures.

Time series $\mathbf{x}_{1:C+H} = [x_1, \dots, x_{C+H}]$, where $C$ denotes the context length and $H$ the forecast horizon, are first normalised using mean scaling, defined as $\tilde{x}_i = x_i / s$ with $s = \frac{1}{C} \sum_{i=1}^{C} |x_i|$. This choice ensures scale-invariance while preserving zero values, which are often semantically significant. The normalised values are subsequently quantised into a fixed number $B$ of discrete bins via a quantisation function $q : \mathbb{R} \rightarrow \{1, \dots, B\}$, where bins are defined by centres $c_1 < \dots < c_B$ and edges $b_1, \dots, b_{B-1}$ with $b_i = \frac{c_i + c_{i+1}}{2}$. Uniform binning is employed to enhance generalisation to unseen distributions. Dequantisation is achieved through a mapping $d(j) = c_j$, enabling reconstruction of real-valued outputs.

The discrete sequence $\mathbf{z}_{1:C+H} = [q(\tilde{x}_1), \dots, q(\tilde{x}_{C+H})]$ is augmented with special tokens, such as \texttt{PAD} for padding and \texttt{EOS} for sequence termination. These tokenised sequences are processed by standard transformer architectures, whose embedding layers are resized to accommodate the new vocabulary $\mathcal{V}_{\text{ts}}$ of size $|\mathcal{V}_{\text{ts}}| = B + 2$.

Training is conducted via maximum likelihood estimation using a categorical cross-entropy loss:
\begin{equation}
   \ell(\bm{\theta}) = -\sum_{h=1}^{H+1} \sum_{i=1}^{|\mathcal{V}_{\text{ts}}|} \mathbf{1}_{(z_{C+h+1} = i)} \log p_{\bm{\theta}}(z_{C+h+1} = i \mid \mathbf{z}_{1:C+h}),
   \label{eq:Chronos-loss}
\end{equation}
where $p_{\bm{\theta}}$ is the predictive categorical distribution over tokens, parameterised by model weights $\bm{\theta}$. Despite the loss function lacking explicit distance-awareness across token bins, empirical results indicate that the model internalises the ordinal structure from the data distribution. This setup enables regression via classification~\cite{torgo1997regression,stewart2023regression}, and allows the model to represent nonparametric, potentially multimodal predictive distributions.

Forecasting is performed by autoregressive sampling of token sequences $\hat{z}_{C+1}, \dots, \hat{z}_{C+H}$ from the model’s output distribution, followed by dequantisation and inverse scaling to recover $\hat{x}_{C+1}, \dots, \hat{x}_{C+H}$. This produces a probabilistic forecast in the form of multiple sampled trajectories.

To mitigate the limited diversity and scale of public time series datasets compared to natural language corpora, the \texttt{Chronos} framework introduces two forms of data augmentation. The first, TSMixup, extends the classic mixup technique~\cite{zhang2017mixup} to time series by generating convex combinations of $k \in \{1, \dots, K\}$ randomly selected sequences, where $k$ and sequence length $l$ are drawn from uniform distributions. The coefficients $\lambda_i$ are sampled from a symmetric Dirichlet distribution, and each series is scaled before mixing to avoid magnitude biases:
\begin{equation}
    \tilde{\mathbf{x}}^{\text{TSMixup}}_{1:l} = \sum_{i=1}^k \lambda_i \tilde{\mathbf{x}}^{(i)}_{1:l}.
\end{equation}

The second augmentation strategy, KernelSynth, generates synthetic time series by sampling from Gaussian Process (GP) priors. A set of primitive kernels \--- linear, periodic, and RBF \--- is randomly composed via binary operations to create composite kernels $\tilde{\kappa}(t, t')$. New sequences are then sampled from the GP prior $\mathcal{GP}(0, \tilde{\kappa})$. This process, inspired by the Automatic Statistician \cite{duvenaud2013structure}, enables the synthesis of diverse patterns that improve the generalisation capability of the forecasting model.

\section{Data Selection and Exploration}
\label{sec:methodology}

Following \cite{huang2019causal}, we conducted a structural causal analysis on the interrelationships among key macroeconomic indicators in the United States \--- namely GDP, inflation, economic growth, and unemployment \--- over the period from 1970 to 2021, using quarterly normalised data. To facilitate comparison and improve the numerical stability of the causal analysis, as recommended by \cite{huang2019causal}, we normalised all variables using standard scaling. %
This transformation preserves the underlying structure and relationships among variables, making it especially suitable for causal discovery in multivariate time series.

We employ four macroeconomic indicators consistent with the causal framework introduced in \cite{huang2019causal}. \textbf{Economic growth (quarterly)} is measured as the percentage change in Gross Domestic Product (GDP) from the previous quarter, adjusted for seasonality and expressed in constant prices. These data are sourced from the Organisation for Economic Co-operation and Development (OECD). \textbf{GDP (billion currency units)} is obtained from the U.S. Bureau of Economic Analysis (BEA) and represents the sum of the gross value added by all resident producers in the economy, plus product taxes minus subsidies, reported in current USD and aggregated quarterly. \textbf{Inflation} is defined as the monthly percent change in the Consumer Price Index (CPI), as reported by the U.S. Bureau of Labor Statistics; we average this value across each quarter to align with the temporal resolution of other variables. Finally, the \textbf{unemployment rate} is the percentage of the labour force actively seeking work but currently unemployed, also reported by the U.S. Bureau of Labor Statistics. Like inflation, unemployment rates are averaged over the three months comprising each quarter. %
\Cref{fig:time-series} depicts the overall time series after the data normalisation.

The model introduced in \cite{huang2019causal} revealed a contemporaneous causal structure wherein both inflation and economic growth directly affect GDP; economic growth, in turn, influences inflation; and unemployment is causally affected by both GDP and inflation. These results, the authors claim \cite{huang2019causal}, align with established economic intuition.

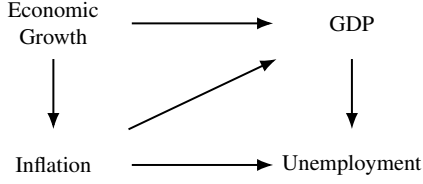
\begin{figure}[bt]
\centering
\resizebox{0.7\linewidth}{!}{%
\begin{tikzpicture}[
  node distance=1cm and 2cm,
  every node/.style={minimum width=2.2cm, minimum height=1cm, align=center},
  arrow/.style={-{Latex}, thick}
]

\node (growth) {Economic\\Growth};
\node[right=of growth] (gdp) {GDP};
\node[below=of gdp] (unempl) {Unemployment};
\node[below=of growth] (infl) {Inflation};

\draw[arrow] (growth) -- (gdp);
\draw[arrow] (gdp) -- (unempl);
\draw[arrow] (growth) -- (infl);
\draw[arrow] (infl) -- (gdp);
\draw[arrow] (infl) -- (unempl);

\end{tikzpicture}
}
\caption{Identified contemporaneous causal relationships between GDP, inflation, economic growth, and unemployment, see \cite[Fig. 4]{huang2019causal}.}
\label{fig:scm_macro}
\vspace{2em}
\end{figure}

\begin{figure}[bt]
    \centering
    \includegraphics[width=\linewidth]{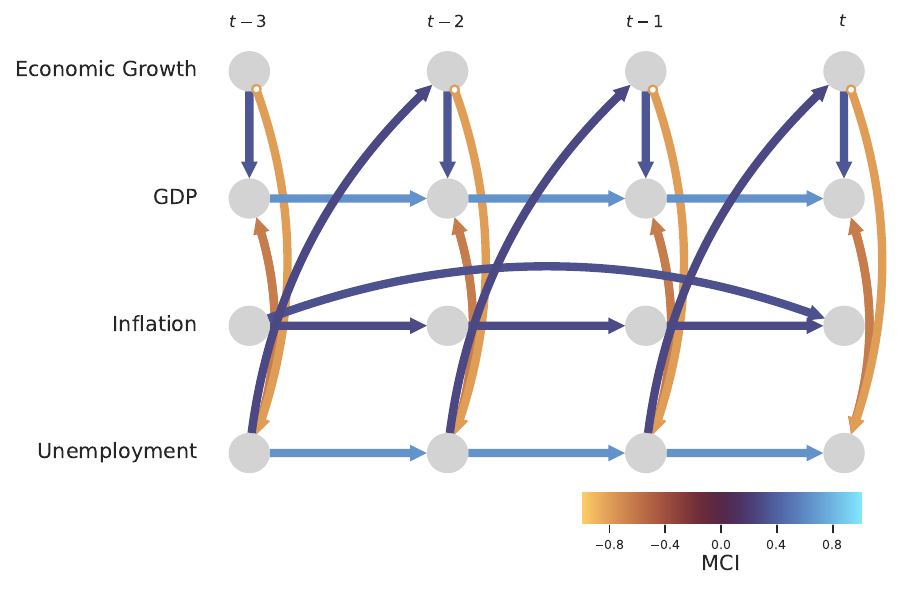}
    \caption{Time-lagged causal graph representation of multivariate time series, visualised across four variables from time steps $t-3$ to $t$. Directed edges indicate statistically significant causal links, with arrow colour denoting the strength and direction of the multivariate conditional independence (MCI) measure computed using the partial correlation test with analytic significance.}
    \label{fig:causal-parcorr}
    \vspace{3em}
\end{figure}

As a data exploration step, we employed the LPCMCI algorithm to uncover time-lagged causal relationships among the standardised macroeconomic indicators, using the partial correlation test with analytic significance evaluation as the conditional independence criterion. 
In this an the following analyses, we set \(\tau_{\text{max}} = 4\), corresponding to a temporal window of one year given the quarterly frequency of the data, to capture seasonal and delayed effects in economic dynamics, where responses to policy changes, market adjustments, or structural shifts may take several quarters to manifest.
To test whether two variables \( X \) and \( Y \) are conditionally independent given a conditioning set \( Z \), we first regress out the linear effects of \( Z \) from both \( X \) and \( Y \) using ordinary least squares, yielding residuals \( \epsilon_X \) and \( \epsilon_Y \) via:
\begin{equation}
X = Z \beta_X + \epsilon_X, \quad Y = Z \beta_Y + \epsilon_Y.
\end{equation}
We then test the residuals for dependence using the Pearson correlation coefficient:
\begin{equation}
\rho(\epsilon_X, \epsilon_Y).
\end{equation}
Under the null hypothesis of conditional independence, the test statistic is assumed to follow a Student's-\emph{t} distribution with \( T - D_Z - 2 \) degrees of freedom, where \( T \) denotes the sample size and \( D_Z \) the dimensionality of the conditioning set \( Z \). %
The results of this analysis are depicted in \Cref{fig:causal-parcorr}.

While LPCMCI with the partial correlation test provides a computationally efficient framework for discovering causal links in multivariate time series, its reliance on linear and Gaussian assumptions can limit its ability to detect non-linear dependencies, leading to unreliable or noisy causal inferences in real-world data. %
As a result, spurious dependencies may emerge from model mis-specification or underfitting, inflating false positives.

\section{Causal Analysis and Zero-Shot Uncertainty-Aware Time Series Prediction}
We instead proceed with an analysis using the Gaussian Process Distance Correlation (GPDC) test as introduced in \Cref{sec:lpcmci}, which adopts a fully nonparametric two-stage procedure, modelling conditional expectations \( f_X(Z) \) and \( f_Y(Z) \) via Gaussian processes that can flexibly capture highly nonlinear relationships. %

\subsection{Causal Analysis with GPDC}
\label{sec:resultsgpdc}

\begin{figure}[tb]
    \centering
    \includegraphics[width=\linewidth]{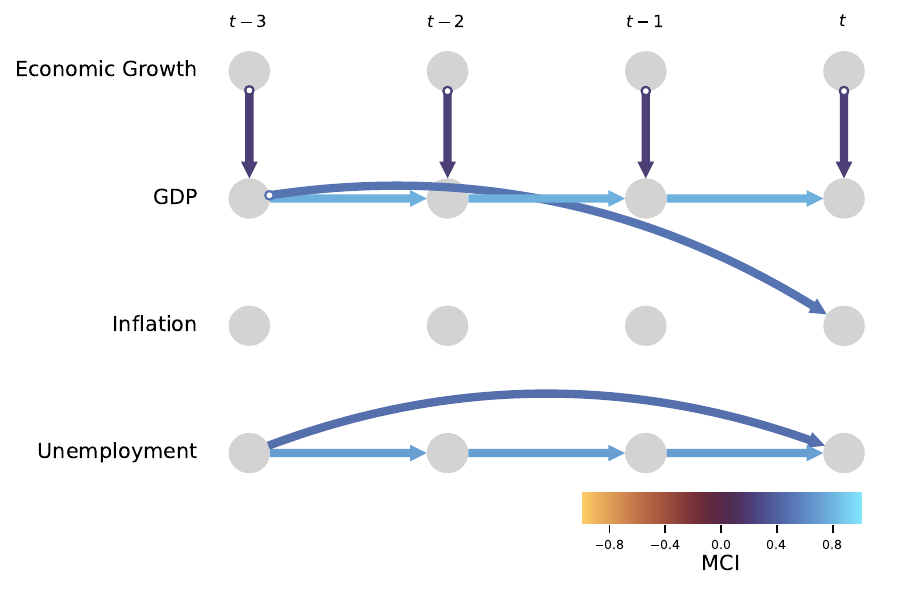}
    \caption{Time-lagged causal graph representation of multivariate time series, visualised across four variables from time steps $t-3$ to $t$. Directed edges indicate statistically significant causal links, with arrow colour denoting the strength and direction of the multivariate conditional independence (MCI) measure computed using GPDC.}
    \label{fig:causal-gpdc}
    \vspace{2em}
\end{figure}

The application of the LPCMCI framework paired with the GPDC test (Figure~\ref{fig:causal-gpdc}) provides a more flexible and expressive lens for detecting temporal causal relationships in macroeconomic time series. Unlike linear partial correlation-based methods, GPDC captures a broader spectrum of dependency structures, including nonlinear and non-monotonic interactions, by combining nonparametric regression via Gaussian processes with a rank-based distance correlation test on residuals. %
The resulting causal graph offers insights that differ significantly from those inferred using more constrained methods.

One result is the strong causal link from Economic Growth to GDP, which persists even under the nonparametric GPDC framework. This finding aligns with theoretical and empirical expectations: economic growth, typically measured through increases in output and productivity, is fundamentally linked to the evolution of gross domestic product over time. The stability and strength of this relationship across model specifications lend credibility to the approach and affirm the structural nature of this dependency. 

More unexpected, however, is the behaviour of inflation. In \Cref{fig:causal-gpdc}, inflation appears to be only loosely connected with the other macroeconomic indicators \--- and even lacks a strong autoregressive component. This weak integration may indicate that the temporal evolution of inflation is governed by latent or exogenous factors not captured in the present model, such as global commodity prices, monetary policy shocks, or anticipatory expectations not directly observable in standard macroeconomic aggregates. In contrast, the variable Unemployment displays a markedly self-referential pattern, with strong quarter-over-quarter dependencies suggesting high temporal inertia. This observation aligns with known frictions in labour markets, such as delayed hiring/firing responses, regulatory constraints, or mismatch between labour supply and demand.

Given the pronounced temporal self-dependence observed in the Unemployment series \--- characterised by robust causal links to its own lagged values across consecutive quarters \--- we investigate the predictive structure of Unemployment using a zero-shot probabilistic forecasting approach based on a large pretrained language model.

\section{Zero-Shot Uncertainty-Aware Time Series Prediction}
\label{sec:chronosprediction}

\begin{figure*}
    \centering
    \includegraphics[width=\linewidth]{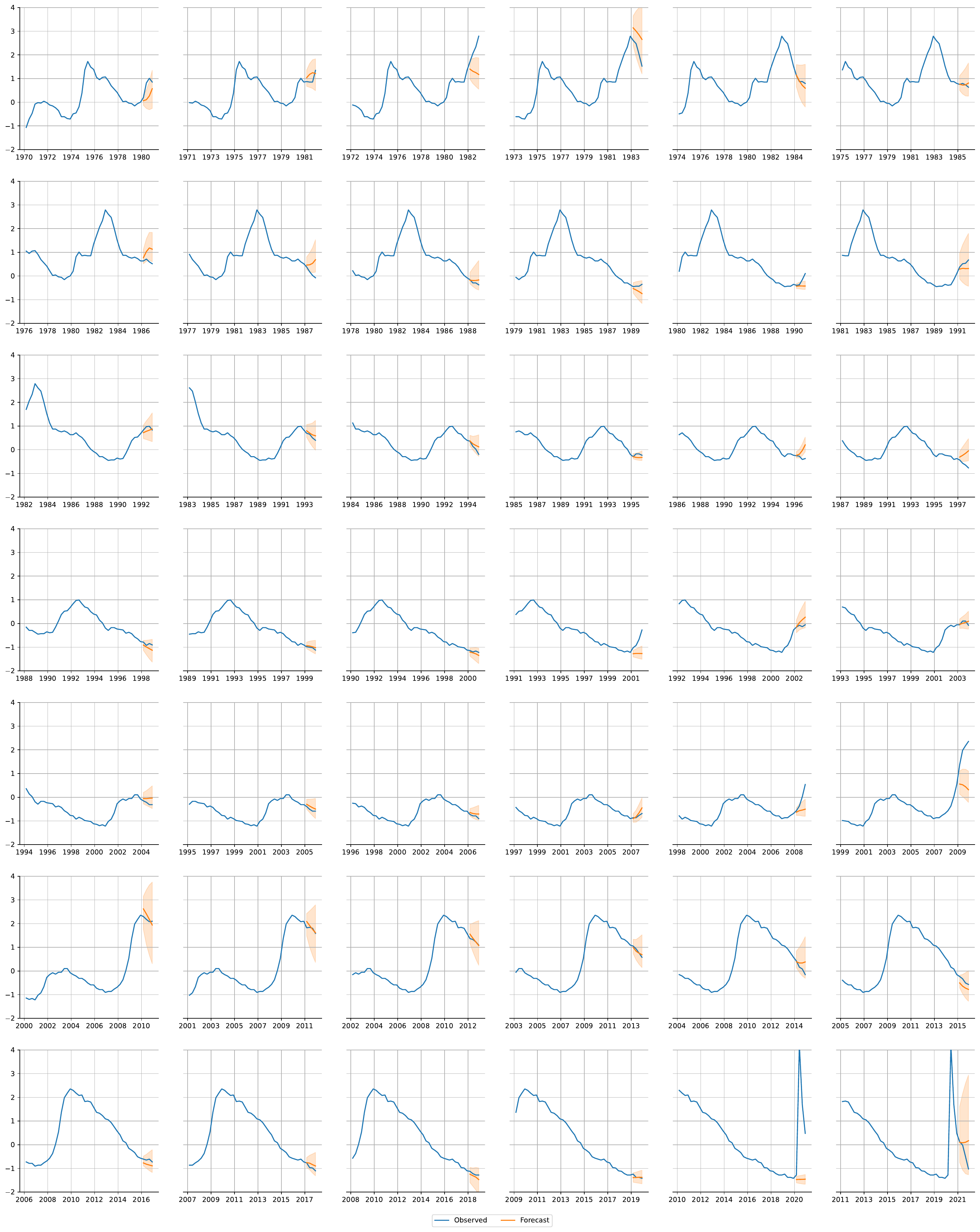}
    \caption{Zero-shot forecast of unemployment using \texttt{Chronos}. Blue denotes the actual values, orange the predicted mean, and the shaded area indicates the 90\% prediction interval.}
    \label{fig:forecast}
\end{figure*}

We employed the \texttt{\texttt{Chronos}} framework introduced in \Cref{sec:Chronos} to perform probabilistic zero-shot forecasting of the Unemployment time series. The model operates on a windowed representation of the data, where a fixed-length context is used to generate probabilistic predictions over a future horizon. In our setup, we prompted the model with ten years of historical quarterly data to forecast the subsequent year \--- \eg using data from 1970 to 1979 (inclusive) to predict values for 1980. The results, depicted in Figure~\ref{fig:forecast}, illustrate the real observations in blue, the model's predicted mean trajectory in orange, and the 90\% prediction interval as a shaded region. The \texttt{Chronos} framework directly produces these intervals, which outputs complete predictive distributions rather than point estimates.

Crucially, this probabilistic treatment allows us to represent the \emph{total predictive uncertainty}, which encompasses both aleatory and epistemic components. Aleatory uncertainty reflects inherent variability in the data \--- such as stochastic fluctuations in unemployment rates due to unforeseen macroeconomic shocks \--- and cannot be reduced by additional observations. Epistemic uncertainty, by contrast, arises from limited knowledge or model uncertainty, and is particularly relevant in zero-shot or distribution-shift settings where the model has not encountered similar patterns during pretraining. While \texttt{Chronos} does not explicitly disentangle these sources, its predictive intervals widen in regions of higher overall uncertainty, which may stem from either or both components. This aligns with recent efforts in developing uncertainty-aware deep learning models \cite{Gal2016Dropout, malinin_Predictiveuncertaintyestimation_18, Sensoy2018, sensoy_UncertaintyAwareDeepClassifiers_20,cerutti_EvidentialReasoningLearning_22, cerutti2022handling}, aimed at improving decision-making reliability in high-stakes domains. %

By inspection of \Cref{fig:forecast}, we observe that the prediction is notably inaccurate for 1982, with the model substantially underestimating the rise in unemployment. This discrepancy arises because the model, trained on data from the preceding decade, could not anticipate the sharp labour market deterioration triggered by the Volcker-induced recession. Similar limitations occur for 2001, 2008--2009, and 2020--2021, corresponding to the dot-com bust, the global financial crisis, and the COVID-19 pandemic, respectively. These events represent structural breaks and distribution shifts that fall outside the historical patterns seen during pretraining, thereby challenging zero-shot forecasting.

Regarding calibration, the 90\% prediction intervals produced by \texttt{Chronos} cover approximately 81\% of the 42 forecasted observations (about 8 violations). While this undercoverage suggests some degree of miscalibration, it is important to note that (i) the evaluation set is small, so sampling variability is high \--- under perfect calibration, the number of violations follows a $\text{Binomial}(42, 0.1)$ distribution, where observing about 8 violations is within two standard deviations of the expected value; and (ii) the largest errors coincide with unprecedented shocks, where both aleatory and epistemic uncertainties are elevated. \texttt{Chronos} does not guarantee exact nominal coverage, but its intervals adaptively widen after shocks, reflecting increased total predictive uncertainty.

From an anomaly detection perspective, this time series is particularly informative when the observed value lies significantly outside the model's predicted distribution. Indeed, if the true observation $x_t$ lies in the tails of the predictive distribution $p(x_t \mid x_{t-10:t-1})$, such that $P(x_t \leq \hat{x}_t) < \alpha$ or $P(x_t \geq \hat{x}_t) > 1 - \alpha$ for a small $\alpha$ (\eg 0.05), then $x_t$ may be flagged as an anomaly. This statistical deviation from the model's credible interval can serve as a principled indicator of structural breaks or regime shifts, making the predictive framework useful for forecasting and real-time anomaly detection in economic surveillance.

\begin{figure}[tb]
    \centering
    \includegraphics[width=\linewidth]{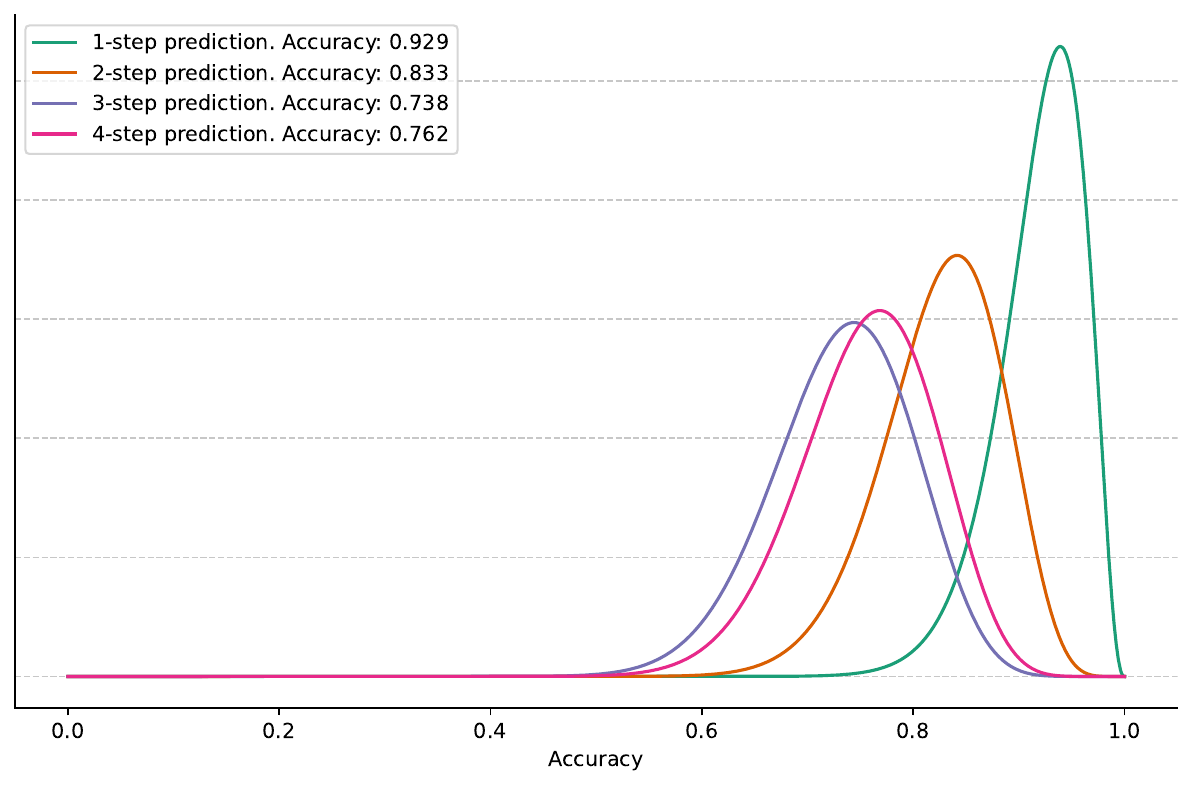}
    \caption{Posterior Beta distributions of the predictive accuracy at one, two, three, and four timesteps ahead, estimated using a Jeffreys prior. The distributions provide a Bayesian characterisation of model performance across increasing forecast horizons, highlighting both the central tendency and associated uncertainty.}
    \label{fig:accuracy-beta}
    \vspace{2em}
\end{figure}

\paragraph{Prediction Accuracy}
To assess the predictive reliability of zero-shot learning, \Cref{fig:accuracy-beta} presents the posterior Beta distributions for the empirical coverage probability at one, two, three, and four timesteps ahead (on quarterly figures). For each horizon, we define a binary indicator that equals 1 if the observed value falls within the 90\% prediction interval produced by \texttt{Chronos}, and 0 otherwise. The total number of ``successes'' (observations inside the interval) is then modelled as a Binomial random variable, and inference is conducted using a Beta distribution with Jeffreys prior, $\text{Beta}(0.5, 0.5)$. This non-informative prior is commonly used in Bayesian inference for proportions, reflecting prior ignorance while ensuring a proper posterior.

Given $k$ correct predictions out of $n$ trials, the posterior distribution becomes $\text{Beta}(k + 0.5, n - k + 0.5)$, which quantifies the uncertainty about the true coverage probability due to the finite sample size. This approach avoids relying solely on a point estimate and instead provides a full posterior distribution, enabling an uncertainty-aware evaluation and a more robust comparison of predictive performance across different forecasting horizons.

Looking at the resulting distributions, we observe that the average predictive accuracy declines as the forecast horizon increases: from 0.929 at one step ahead, to 0.833, 0.738, and 0.762 at four steps ahead. While the four-step value is slightly higher than the three-step value, this difference is small and may reflect sampling variability rather than a systematic effect. One possible explanation, which would require further investigation, is that the model’s architecture allows it to smooth over temporal dependencies at longer horizons, partially mitigating short-term shocks. This interpretation is also supported by the causal structure identified in \Cref{fig:causal-gpdc}, highlighting a significant dependency between lagged values at $t-3$ and the current state at $t$.

\begin{figure}[tb]
    \centering
    \includegraphics[width=\linewidth]{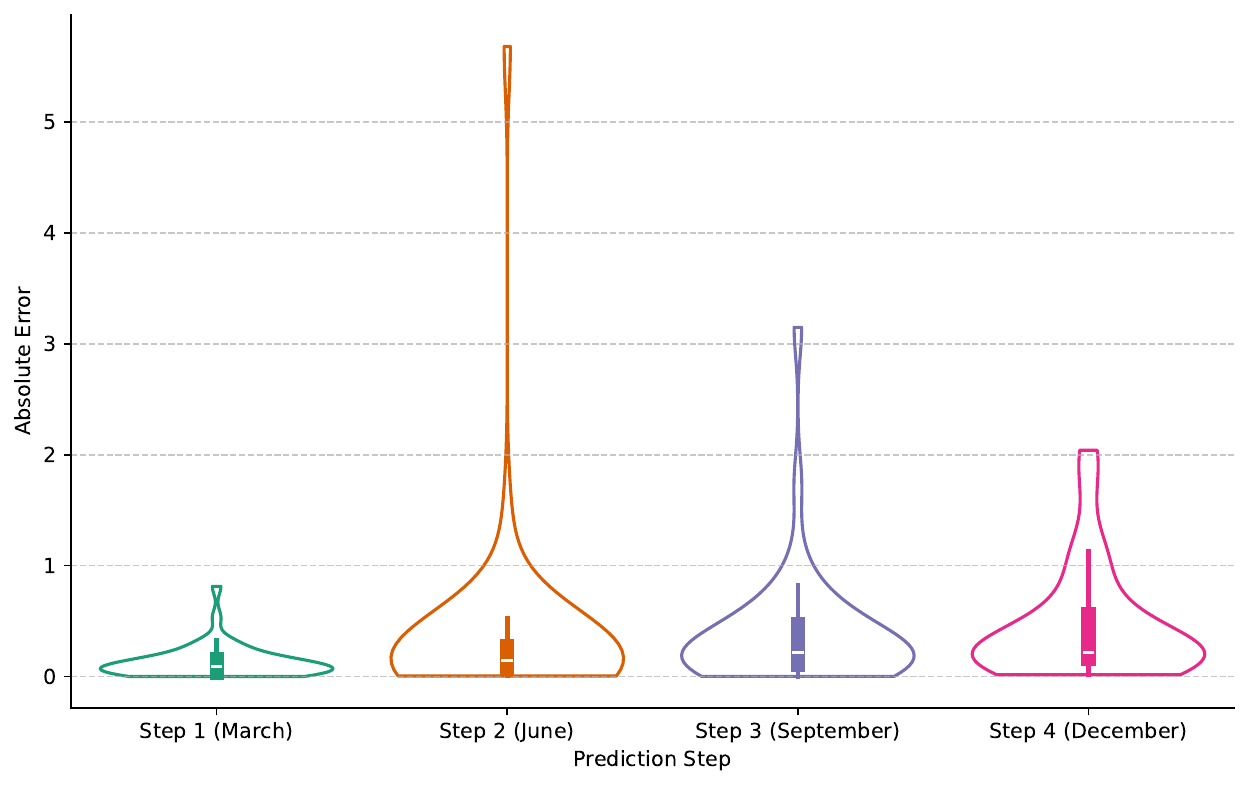}
    \caption{Violin plots of the absolute prediction error at one, two, three, and four timesteps ahead.}
    \label{fig:abserror}
    \vspace{2em}
\end{figure}

\paragraph{Absolute Error Distribution}
\Cref{fig:abserror} presents the distribution of absolute errors at one, two, three, and four timesteps ahead using violin plots. This visualisation provides a non-parametric representation of the full distribution of the absolute deviation \(|y_t - \hat{y}_t|\), allowing for a nuanced comparison of predictive performance across different forecast horizons. The choice of absolute error rather than squared error is motivated by its robustness to outliers.%

As expected, the sample average of the absolute error increases with the forecast horizon, reflecting the compounding uncertainty over time. The violin plot for the two-step horizon appears to exhibit heavier tails; however, this impression is driven by a single extreme observation (the second-to-last point of \Cref{fig:forecast}) rather than a systematic pattern. Consequently, we refrain from attributing a higher intrinsic risk of large deviations to the two-step forecast. Overall, the distributions highlight that while most errors remain moderate, occasional outliers can occur, particularly during periods of structural change.

\begin{figure}[tb]
    \centering
    \includegraphics[width=\linewidth]{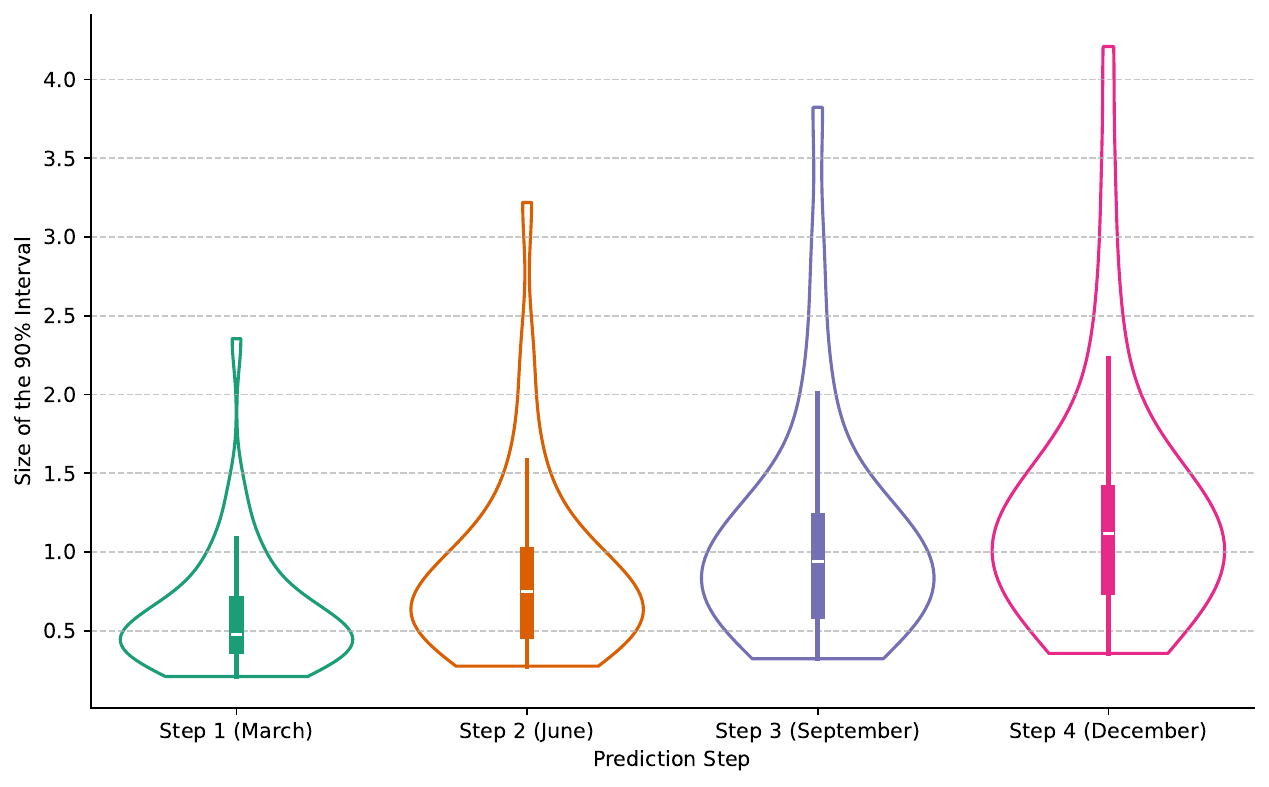}
    \caption{Distribution of the size of the 90\% prediction interval at one, two, three, and four timesteps ahead.}
    \label{fig:size90interval}
    \vspace{2em}
\end{figure}

\paragraph{Distribution of Prediction Intervals}
\Cref{fig:size90interval} displays the size of the 90\% prediction interval at one, two, three, and four timesteps ahead. Each interval size is computed as the difference between the upper and lower bounds of the model's probabilistic forecast at a given time point, aggregated over the test set. This figure provides a direct visualisation of the model’s quantified uncertainty.%

As expected, the average size of the 90\% prediction interval increases with the forecast horizon, reflecting greater total predictive uncertainty over longer time spans. This widening is a desirable property in uncertainty-aware models, as it indicates recognition of diminishing predictive precision with temporal distance.

\section{Conclusion and Future Work}
This study demonstrates the complementary strengths of structural causal models and uncertainty-aware forecasting in analysing macroeconomic time series. By applying the LPCMCI framework combined with the GPDC conditional independence test, we uncover meaningful causal relations among GDP, economic growth, inflation, and unemployment \--- revealing, for instance, the autoregressive dynamics of unemployment. Building on this, we employ a zero-shot probabilistic forecasting approach using \texttt{Chronos} to model unemployment, showcasing the feasibility of high-accuracy, uncertainty-calibrated predictions without retraining.
Our results underscore the potential of integrating causal discovery with uncertainty-aware forecasting to support robust data interpretation and policy design. The capacity to flag anomalies through distributional deviation further enhances the practical utility of such models.%

Future work will extend this framework beyond single-variable forecasting to investigate multivariate predictive dependencies. In particular, we aim to characterise joint predictive distributions across functionally interrelated indicators, exploring how uncertainty propagates through the system. This direction will require new modelling strategies that balance computational efficiency with expressive capacity, potentially integrating probabilistic graphical models with large pre-trained time series forecasters.

\begin{ack}
We thank the anonymous reviewers for their insightful and constructive feedback, which greatly improved the clarity and rigour of our analysis.
The work was partially supported by the European Office of Aerospace Research \& Development and the Air Force Office of Scientific Research under award numbers FA8655-22-1-7017 and FA8655-25-1-7067, and by the US DEVCOM Army Research Laboratory (ARL) under Cooperative Agreement \#W911NF2220243. Any opinions, findings, and conclusions or recommendations expressed in this material are those of the author(s) and do not necessarily reflect the views of the United States government.
This work was supported by the EU NEXTGENERATIONEU program within the PNRR Future Artificial Intelligence – FAIR project (PE0000013, CUP H23C22000860006), Objective 10: Abstract Argumentation for Knowledge Representation and Reasoning, specifically by the project Argumentation for Informed Decisions with Applications to Energy Consumption in Computing – AIDECC (CUP D53C24000530001).
GPT-4o and GPT-5 were used for text refinement and language editing.%

\end{ack}

\bibliography{mybibfile,biblio,morebib}

\end{document}